\documentclass{article} 
\usepackage{iclr2027_conference,times}

\usepackage{amsmath,amsfonts,bm}

\def\eqref#1{equation~\ref{#1}}

\def\1{\bm{1}}

\DeclareMathAlphabet{\mathsfit}{\encodingdefault}{\sfdefault}{m}{sl}
\SetMathAlphabet{\mathsfit}{bold}{\encodingdefault}{\sfdefault}{bx}{n}

\usepackage{hyperref}
\usepackage{url}

\usepackage{graphicx}
\usepackage{subcaption}
\usepackage{booktabs}
\usepackage{colortbl}
\usepackage{arydshln}
\usepackage{multirow}
\usepackage{amssymb}
\usepackage{wrapfig}
\title{Quality Determines Direction, Length Shapes Magnitude: Length Control for Open-Ended Reinforcement Learning}

\author{
Zijun Weng$^{1,2,\dagger}$,
Zhongan Bi$^{3,\dagger}$,
Xuanang Gao$^{4,\dagger}$,
Xiaohui Hu$^{2}$,
Shuangyong Song$^{2}$,
\\[-1pt]
Yongxiang Li$^{2}$,
Kaidong Yu$^{2,*}$,
Xuanjing Huang$^{1,*}$
}

\begin{document}

\maketitle
\begingroup
\renewcommand{\thefootnote}{}
\footnotetext{
\scriptsize
$^{1}$Fudan University;
$^{2}$Xingchen AGI Lab, China Telecom Artificial Intelligence Technology (Beijing) Co., Ltd.;
$^{3}$Zhejiang University;
$^{4}$Shanghai Jiao Tong University.
$^{\dagger}$Work performed while at Xingchen AGI Lab.
$^{*}$Corresponding authors.
}
\endgroup

\begin{abstract}
Reinforcement learning (RL) changes not only what language models say, but also how much they say, often increasing response length at the cost of token efficiency. Controlling this length growth is particularly challenging in open-ended RL because \textbf{(i)} response length is entangled with quality, \textbf{(ii)} open-ended tasks lack a natural success boundary for deciding when efficiency should be prioritized, and \textbf{(iii)} dense, graded rewards often yield small within-group quality margins, making quality-induced advantages especially sensitive to reward-level length shaping, which can perturb their magnitudes and even reverse their signs.
We therefore adopt an asymmetric principle: \textbf{quality should determine the direction of reinforcement, while length should only shape its magnitude}. We instantiate this principle with \textbf{Quality-Gated Length Advantage Shaping (QGLAS)}, which first computes advantages from quality rewards alone, then adds bounded bonuses only to shorter positive-advantage responses, leaving all other advantages unchanged. The bonus strength is further adapted to within-group quality separation, allowing conciseness to matter more when quality-favored responses are similar and less when their quality differences are clear.
Across different model families, open-ended benchmarks, and reward sources, QGLAS consistently achieves a stronger quality--length trade-off than representative baselines. At approximately 30\% compression, QGLAS retains 98.4--102.0\% of the macro-average quality gains achieved by quality-only RL over the base
model, compared with 68.3--75.5\% for these baselines at comparable
compression.
\end{abstract}

\begin{wrapfigure}{r}{0.50\textwidth}
    \centering
    \vspace{-8pt}
    \includegraphics[width=\linewidth]{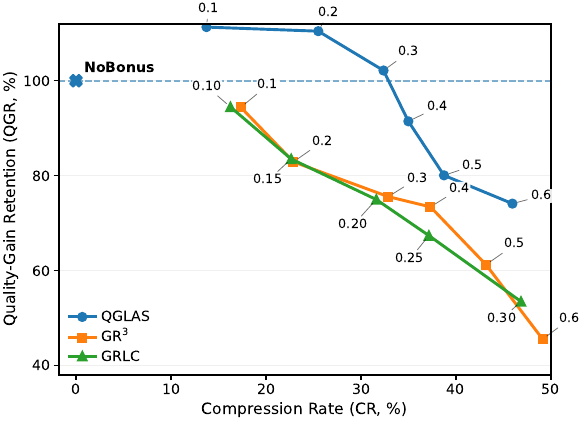}
    \caption{
    \textbf{Quality--length trade-off on Qwen3-4B.}
    QGLAS retains substantially more quality gain than others at comparable compression.
    }
    \label{fig:pareto}
    \vspace{-10pt}
\end{wrapfigure}

\section{Introduction}
\label{sec:introduction}

Reinforcement learning (RL) changes not only what language models say,
but also how much they say. On open-ended tasks such as instruction
following, dialogue, decision making, and creative generation, quality
gains often accompany substantially longer responses
\citep{chen2024odin}. Additional tokens can improve coverage and
justification, but longer responses may also receive higher reward or
evaluation scores without commensurate quality gains
\citep{bu-etal-2025-beyond,dubois2024alpacaeval,nohara2026on}.
RL can therefore exploit verbosity as an optimization shortcut, yet
indiscriminate length penalties may suppress useful content.
Effective length control must reduce low-value generation while
preserving genuine quality gains.

Open-ended RL makes this problem challenging because \textbf{(i)} length is entangled with quality: additional tokens may represent either redundancy or useful reasoning and detail; \textbf{(ii)} graded feedback lacks a natural success boundary for prioritizing efficiency, and absolute reward values need not have comparable meanings across prompts or reward sources; and \textbf{(iii)} dense feedback can yield small within-group quality margins, making relative learning signals sensitive to length-dependent perturbations. Selecting responses for length control and calibrating its strength therefore require attention to the local quality structure.

Many RL-based length-control methods are developed for
verifiable-reward settings, where correctness or group-level success
statistics guide the balance between task performance and efficiency
\citep{liu2026laser,pranjal2025l1,peng2026thinkdenselongdynamic}.
Recent methods, including Group Relative Reward Rescaling (GR$^3$)   \citep{li2026gr3} and Group Relative Length Control (GRLC) \citep{nvidia2025grlc}, support continuous feedback
and incorporate quality-dependent safeguards.
However, they introduce length into the rewards used to compute relative advantages.
Consequently, reward-level shaping can alter not only the reinforcement polarity of directly targeted responses, but also the original quality-induced advantages of responses that receive no length adjustment at all.

We examine this effect in a controlled analysis.
Applying GR$^3$ with the same coefficient to identical stored rollout
groups yields a macro-average sign-reversal rate of $11.73\%$ under
dense rewards versus $0.30\%$ averaged over four binarized
variants---a nearly $40\times$ difference.
Although this comparison does not reproduce RLVR training,
it shows that support for continuous feedback alone does not ensure
preservation of the reinforcement decisions induced by quality.

These observations motivate an asymmetric principle:
\emph{quality should determine the direction of reinforcement,
while length should only shape its magnitude}.
We express this requirement as \emph{quality-polarity invariance}:
length shaping must preserve the sign of every advantage computed
from the original quality rewards.
Within this constraint, conciseness may still modulate the relative
reinforcement strength among quality-favored responses.

We propose \textbf{Quality-Gated Length Advantage Shaping (QGLAS)}
to enforce this requirement by construction.
QGLAS computes advantages from quality rewards alone, then adds
bounded bonuses only to shorter positive-advantage responses.
Advantage-level shaping leaves untargeted advantages unchanged,
while positive-only, one-sided bonuses preserve every advantage sign.
The shaping strength adapts to the overall reward spread and quality
separation among favored responses, giving conciseness more weight
when their rewards are similar and less when they are clearly
separated.

On Qwen3-4B and GLM-4.7-Flash, QGLAS achieves a stronger
quality--length trade-off than GR$^3$ and GRLC.
At approximately \textbf{30\%} compression relative to quality-only
RL, it retains \textbf{98.4--102.0\%} of that policy's macro-average
quality gain over the base model, compared with
\textbf{68.3--75.5\%} for these baselines at comparable compression.
Repeated runs and alternative reward sources further support its
robustness.
A fixed-strength ablation also preserves every advantage sign but
retains substantially less quality, showing the benefit of adapting
shaping strength within the same constraint.
Our contributions are threefold:
\begin{itemize}
    \item We empirically diagnose advantage sign reversals under
    reward-level length shaping, showing substantially greater
    sensitivity under dense feedback than under controlled
    binarizations of the same quality rewards.

    \item We introduce QGLAS, which isolates length shaping to explicitly targeted responses,
    preserves quality-induced reinforcement polarity by construction, and adapts shaping
    strength to the local reward structure.

    \item We demonstrate improved quality--length trade-offs
    across policy models and reward sources, with ablations
    supporting complementary benefits from polarity-preserving
    constraints and adaptive shaping strength.
\end{itemize}

\section{Related Work}
\label{sec:related_work}

\paragraph{Length bias and response efficiency.}
Excessive response length has been studied at multiple stages of the alignment
pipeline. Length-controlled evaluation reduces verbosity-related confounding
\citep{dubois2024alpacaeval,hu2025explaining}, while reward-modeling approaches
seek to separate genuine quality from superficial preferences for longer
responses \citep{shen2023loose,liu2025rrm}. A complementary line intervenes
directly during RL through length-dependent optimization signals
\citep{li-etal-2026-leash,han2026jet,xiang2025alp,liu-etal-2026-tokens}.
Our work follows this RL-time setting, focusing on how efficiency preferences
should interact with graded quality feedback.

\paragraph{Length control with verifiable rewards.}
Many RL-based length-control methods are developed for reinforcement learning
with verifiable rewards (RLVR). They use correctness, solve rate,
or estimated difficulty to determine which responses to encourage toward
conciseness and how strongly
\citep{liu2026laser,yuan-etal-2026-shorten,xiang2025alp,li-etal-2026-leash,liu2026laconic}.
For example, Dynamic Decoupled Conditional Advantage (DDCA)
\citep{peng2026thinkdenselongdynamic} computes length advantages within the
correct-response subset and scales their strength by the group pass rate.

These designs exploit two properties of binary feedback: correctness provides
an eligibility criterion, and successful responses share the same
task reward, making length a natural secondary preference within that subset.
These foundations do not transfer directly to graded open-ended feedback,
where no success threshold exists and quality-favored responses
may still differ meaningfully.
Length control must therefore determine both \emph{which} responses are
eligible for efficiency optimization and \emph{how strongly} conciseness
should matter relative to their remaining quality differences.

\paragraph{Length control with continuous quality rewards.}
Representative methods also support length control beyond binary correctness.
Group Relative Length Control (GRLC), introduced in the Nemotron~3 Nano
technical report \citep{nvidia2025grlc}, adds group-relative length adjustments
and quality-gated conciseness bonuses to the quality reward.
Group Relative Reward Rescaling (GR$^3$) \citep{li2026gr3}, our closest
baseline, uses multiplicative reward rescaling, group-relative length
normalization, and advantage-aware calibration to protect a representative
maximum-reward response with group-average length.
Both methods support continuous feedback and incorporate safeguards against
indiscriminate length reduction.

However, these safeguards do not fully address the challenges posed by
graded quality differences.
Both methods incorporate length into rewards before relative advantages are
computed, without requiring every response to preserve the reinforcement
polarity induced by quality alone.
This distinction matters when dense feedback yields small within-group
quality margins.
Moreover, their quality-dependent gating and calibration do not explicitly
adapt shaping strength to the quality separation among positive-advantage
responses within the current group.

Our work treats the quality-induced advantage as the primary learning signal:
quality determines which responses are reinforced or suppressed, while
length modulates reinforcement strength through bonuses restricted to
shorter positive-advantage responses.
We further adapt bonus strength to the overall reward spread and quality
separation within this favored subset, combining polarity preservation
with quality-adaptive length control.
Conciseness thus receives greater weight when quality-favored responses
differ little.

\section{Quality-Gated Length Advantage Shaping}
\label{sec:method}

We propose \textbf{Quality-Gated Length Advantage Shaping (QGLAS)}
for controlling response length in open-ended reinforcement learning.
Our key principle is asymmetric:
\emph{quality determines the reinforcement polarity, while length only
shapes its magnitude}.
Accordingly, QGLAS preserves whether a response is reinforced or
suppressed according to the quality signal, while allowing conciseness
to modulate the relative reinforcement strength among quality-favored
responses.
QGLAS realizes this principle through three design choices:
isolating the length signal from non-targeted responses,
restricting length shaping to quality-favored responses,
and adapting its strength to the local quality structure.

For a rollout group $g$, let $A_i^q$ denote the quality-induced advantage
of response $i$, computed by the underlying group-relative RL algorithm
from the original quality rewards before introducing any length-dependent
signal. QGLAS modifies this advantage as
\begin{equation}
\label{eq:qglas}
\widetilde{A}_i
=
A_i^q
+
\lambda_g h_i .
\end{equation}
Here, $h_i \in [0,1]$ is a response-level length-shaping coefficient
that determines whether response $i$ receives a conciseness bonus and,
if so, by how much. The group-level scale $\lambda_g \geq 0$ controls
how strongly conciseness is allowed to influence the current rollout
group. As defined below, $h_i$ is nonzero only for shorter responses
whose quality-induced advantage is already positive.

\subsection{Isolating the Length Signal}
\label{sec:quality_direction}

QGLAS determines the reinforcement polarity from $A_i^q$ before
introducing any length signal. To see why this matters, consider
group-wise standardized advantages \citep{guo2025grpo},
\begin{equation}
A_i^q=\frac{r_i-\bar r}{\sigma_r}.
\end{equation}
If reward-level length shaping instead perturbs the reward as
$r_i' = r_i+\delta_i$, the resulting advantage is
\begin{equation}
A_i'
=
\frac{(r_i-\bar r)+(\delta_i-\bar\delta)}
     {\sigma_{r+\delta}}.
\end{equation}
Thus, length shaping affects responses even when they receive no direct
perturbation: for $\delta_i=0$, their centered reward is shifted by
$-\bar\delta$. In particular, when $\bar\delta>0$, a positive quality
advantage can reverse sign if $0<r_i-\bar r<\bar\delta$, a risk that is
amplified when dense rewards yield small within-group quality margins.

QGLAS instead shapes the already established quality advantage.
Responses with $h_i=0$ therefore remain exactly unchanged,
$\widetilde A_i=A_i^q$, isolating the length signal to explicitly
targeted responses.

\subsection{Restricting the Scope of Length Shaping}
\label{sec:length_shaping}

QGLAS restricts length shaping through two complementary constraints:
\emph{positive-only gating} and \emph{one-sided shaping}.
Let
$\mathcal{P}=\{i:A_i^q>0\}$
denote the quality-favored responses.
If $\mathcal{P}=\varnothing$, we skip length shaping for the group.
Otherwise, we define a prompt-adaptive reference length and response-level
shaping coefficient as
\begin{equation}
\label{eq:length_gain}
L_{\mathrm{ref}}
=
\frac{1}{|\mathcal{P}|}
\sum_{j\in\mathcal{P}} L_j,
\qquad
h_i
=
\mathbf{1}[i\in\mathcal{P}]
\frac{
\operatorname{clip}\!\left(
\frac{L_{\mathrm{ref}}-L_i}
     {L_{\mathrm{ref}}+\epsilon},
0,c
\right)}
{c}.
\end{equation}
Here, $\epsilon>0$ is a numerical stabilizer, while $c\in(0,1]$
sets the saturation threshold for relative shortening: $h_i$ increases
with relative shortening up to $c$ and is capped at $1$ thereafter.
Thus, $h_i\in[0,1]$ and $h_i>0$ only for quality-favored responses
shorter than $L_{\mathrm{ref}}$.
Positive-only gating prevents brevity from overriding an unfavorable
quality signal, while one-sided shaping avoids penalizing longer
quality-favored responses.
Using only $\mathcal{P}$ to define $L_{\mathrm{ref}}$ also prevents short,
quality-disfavored responses from setting the efficiency target.
Consequently,
\begin{equation}
\operatorname{sign}(\widetilde{A}_i)
=
\operatorname{sign}(A_i^q)
\qquad \forall i,
\end{equation}
so length shaping preserves the original quality-induced reinforcement
polarity.

\subsection{Adapting the Shaping Strength}
\label{sec:adaptive_bonus}

We next define the group-level shaping strength $\lambda_g$.
A fixed shaping strength is poorly calibrated for dense rewards because
their scale and local quality separation can vary substantially across
rollout groups. We define
\begin{equation}
\label{eq:reward_spans}
s_{\mathrm{all}}
=
Q_{100}(r)-Q_{25}(r),
\qquad
s_+
=
\max_{j\in\mathcal{P}} r_j
-
\min_{j\in\mathcal{P}} r_j,
\end{equation}
where $Q_p(r)$ denotes the $p$-th percentile of the group rewards.
Here, $s_{\rm all}$ captures the overall reward scale, while $s_+$ measures the quality separation among quality-favored responses. We use the lower quartile rather than the minimum in $s_{\rm all}$ to make the scale estimate less sensitive to anomalously low-reward responses. This robustness is particularly useful for estimating the overall group scale, whereas $s_+$ is computed only within the quality-favored subset.

We adapt the shaping strength according to the relative separation among
quality-favored responses:
\begin{equation}
\label{eq:adaptive_beta}
\beta_g
=
\beta_{\min}
+
(\beta_{\max}-\beta_{\min})
\left[
1-
\operatorname{clip}\!\left(
\frac{s_+}{s_{\mathrm{all}}+\epsilon},
0,1
\right)
\right].
\end{equation}
When quality-favored responses are weakly separated relative to the
overall reward scale, $\beta_g$ approaches $\beta_{\max}$, allowing
conciseness to act as a stronger secondary preference. As their quality
separation increases, $\beta_g$ decreases toward $\beta_{\min}$, allowing
the original quality differences to dominate.
Under group-wise reward standardization, we define the effective shaping
scale as
\begin{equation}
\label{eq:lambda}
\lambda_g
=
\frac{s_{\mathrm{all}}}{\sigma_r}\beta_g,
\end{equation}
where $\sigma_r$ is computed from the original quality rewards. This
places the length bonus on the same scale as the quality advantage
$A_i^q$.
When $\epsilon$ is negligible and clipping is inactive, the effective
shaping scale can be written approximately as
\begin{equation}
\label{eq:effective_bonus_scale}
\lambda_g
\approx
\frac{
\beta_{\max}s_{\mathrm{all}}
-
(\beta_{\max}-\beta_{\min})s_+
}{
\sigma_r
}.
\end{equation}
This decomposition makes the adaptation explicit: the shaping scale
increases with the overall reward variation $s_{\mathrm{all}}$, but
decreases with the quality separation $s_+$ among quality-favored
responses. Thus, conciseness matters more when favored responses are
difficult to distinguish by quality, and less when quality already
separates them clearly.

If the underlying advantage estimator does not use group-wise
standard-deviation normalization, the corresponding normalization is
omitted, giving $\lambda_g=s_{\mathrm{all}}\beta_g$.
Thus, QGLAS follows the scale convention of the underlying quality
advantage instead of introducing a separate normalization scheme.

\section{Experiments}
\label{sec:experiments}

We evaluate QGLAS's quality--length trade-off, examine advantage sign
reversals under dense feedback and their impact on downstream performance.
We then disentangle structural constraints from adaptive scaling and test
generalization across reward sources.

\subsection{Experimental Setup}
\label{sec:experimental_setup}

\paragraph{Models and training.}
We conduct experiments with Qwen3-4B \citep{qwen3} and GLM-4.7-Flash (30B-A3B) \citep{glm4.7}, covering
different model families and dense and mixture-of-experts architectures.
Training uses the 13K open-ended prompts introduced
by \citet{weng2026prompt}, spanning instruction following, writing, and
decision support. All methods are optimized with GSPO \citep{zheng2025gspo} for
1,000 steps, with 16 rollouts sampled per prompt. Unless otherwise
stated, our main experiments use Skywork-Reward-V2-Llama-3.1-8B
\citep{liu2026skywork} as the quality reward model.

\paragraph{Baselines and control strengths.}
For each policy model, we compare the base model, quality-only RL
(NoBonus), QGLAS, GR$^3$, and GRLC. The latter two are representative reward-level
length-control methods applicable to continuous quality feedback.
All RL methods use the same training data, quality reward, and
optimization budget within each comparison.

GR$^3$ applies multiplicative reward shaping and requires non-negative
rewards to preserve its intended length-control direction. Following its
original RLHF setting, we therefore apply the same reference-based
sigmoid shaping (Preference As Reward; PAR \citep{jiang2024par}) to the raw Skywork scores,
and use PAR for all methods in these comparisons.

For the quality--length sweeps, we only vary each method's control strength. For QGLAS, we sweep
$\beta_{\min}$ with $\beta_{\max}=2\beta_{\min}$; for GR$^3$, we sweep
$\alpha$; for GRLC, we jointly sweep $(\lambda,\beta)$ with
$\beta=\lambda$. Detailed baseline configurations are provided in
Appendix~\ref{app:experimental_details}.

\paragraph{Evaluation.}
We evaluate instruction following on IFBench \citep{pyatkin2026ifbench},
challenging open-ended dialogue on the Hard Prompts subset of
Arena-Hard-v2, and creative writing on its Creative Writing subset
\citep{li2024arena}. We follow the official evaluation protocol for
each benchmark and use GPT-4.1 as the judge model when judge-based
evaluation is required. We report the quality score and average response
length for each benchmark, together with their macro averages. Response
length counts all generated tokens, including thinking tokens and answer tokens.

To summarize the overall quality--length trade-off, we report
quality-gain retention (\(\mathrm{QGR}\)) and compression rate
(\(\mathrm{CR}\)):
\begin{equation}
\mathrm{QGR}(m)
=
\frac{S_m-S_{\mathrm{Base}}}
     {S_{\mathrm{NB}}-S_{\mathrm{Base}}}
\times 100\%,
\qquad
\mathrm{CR}(m)
=
\left(
1-\frac{L_m}{L_{\mathrm{NB}}}
\right)
\times 100\%.
\label{eq:relative_metrics}
\end{equation}
Here, $S_m$ and $L_m$ denote the macro-average quality score and response
length of method $m$, respectively, and $\mathrm{NB}$ denotes
NoBonus. QGR measures the fraction of the quality improvement
from quality-only RL over the base model that is retained after
introducing length control, with $100\%$ indicating full retention.
CR measures the reduction in response length relative to
NoBonus, with larger values indicating stronger compression.

\begin{figure*}[t]
\centering

\begin{subfigure}[t]{0.32\textwidth}
    \centering
    \includegraphics[width=\textwidth]{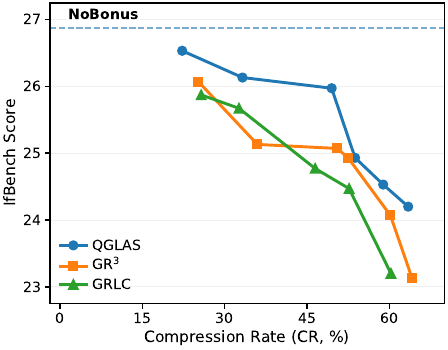}
    \caption{IFBench.}
    \label{fig:pareto_ifbench}
\end{subfigure}
\hfill
\begin{subfigure}[t]{0.32\textwidth}
    \centering
    \includegraphics[width=\textwidth]{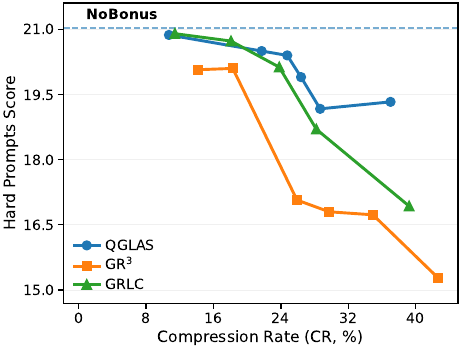}
    \caption{Hard Prompts.}
    \label{fig:pareto_hard}
\end{subfigure}
\hfill
\begin{subfigure}[t]{0.32\textwidth}
    \centering
    \includegraphics[width=\textwidth]{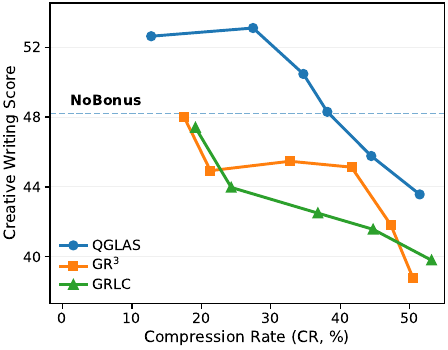}
    \caption{Creative Writing.}
    \label{fig:pareto_creative}
\end{subfigure}

\caption{
Benchmark-level quality--length trade-offs on Qwen3-4B.
Each panel plots benchmark score against benchmark-specific compression
rate (CR) relative to NoBonus across evaluated control strengths.
Higher scores and higher compression are preferred.
}
\label{fig:benchmark_pareto}

\end{figure*}

\subsection{Quality--Length Trade-offs}
\label{sec:main_results}

\paragraph{Trade-offs across control strengths.}
We examine the quality--length trade-off across all evaluated control
strengths. Figure~\ref{fig:pareto} plots QGR against CR for QGLAS,
GR$^3$, and GRLC on Qwen3-4B, while
Figure~\ref{fig:benchmark_pareto} shows the corresponding
benchmark-level trade-offs.

Across the overlapping compression range, QGLAS consistently achieves
higher QGR than GR$^3$ and GRLC at comparable compression.
QGLAS retains essentially the full macro-average quality improvement
of quality-only RL over the base model at compression rates up to
roughly 32\%. In contrast, all evaluated GR$^3$ and GRLC operating
points sacrifice part of this gain, with degradation increasing as
compression increases. The benchmark-level curves show that this
advantage is not driven by a single benchmark. The consistent
separation across the sweeps indicates that QGLAS's advantage is not
specific to a single coefficient.

\begin{table*}[t]
\centering
\caption{
Main results at approximately matched compression.
$\Delta$\textbf{Score} denotes the score change relative to quality-only RL.
\textbf{QGR} and CR are computed from the unrounded macro averages.
Best quality results among length-controlled methods are shown in \textbf{bold}.
}
\label{tab:main_results}

\resizebox{\textwidth}{!}{%
    \begin{tabular}{lrrrrrrrrrr}
        \toprule
        \textbf{Method}
            & \multicolumn{2}{c}{\textbf{IFBench}}
            & \multicolumn{2}{c}{\textbf{Hard Prompts}}
            & \multicolumn{2}{c}{\textbf{Creative Writing}}
            & \multicolumn{2}{c}{\textbf{Macro Average}}
            & \multicolumn{2}{c}{\textbf{Relative Metrics}} \\
        \cmidrule(r){2-3}
        \cmidrule(r){4-5}
        \cmidrule(r){6-7}
        \cmidrule(r){8-9}
        \cmidrule(r){10-11}

            & \textbf{Score}$\uparrow$
            & $\Delta$\textbf{Score}$\uparrow$
            & \textbf{Score}$\uparrow$
            & $\Delta$\textbf{Score}$\uparrow$
            & \textbf{Score}$\uparrow$
            & $\Delta$\textbf{Score}$\uparrow$
            & \textbf{Score}$\uparrow$
            & \#Tok.$\downarrow$
            & \textbf{QGR}$\uparrow$
            & CR$\uparrow$ \\
        \midrule

        \multicolumn{11}{l}{\textbf{\textit{Qwen3-4B}}} \\

        Base
            & 29.22 & --
            & 15.63 & --
            & 16.53 & --
            & 20.46 & 3792
            & 0.0\% & 9.9\% \\

        NoBonus
            & 26.89 & 0.00
            & 21.03 & 0.00
            & 48.20 & 0.00
            & 32.04 & 4207
            & 100.0\% & 0.0\% \\

        \hdashline
        \noalign{\vskip 1mm}

        GR$^3$
            & 25.07 & -1.82
            & 17.07 & -3.96
            & 45.47 & -2.73
            & 29.20 & 2824
            & 75.5\% & 32.9\% \\

        GRLC
            & 24.78 & -2.11
            & 20.13 & -0.90
            & 42.50 & -5.70
            & 29.14 & 2877
            & 74.9\% & 31.6\% \\

        \rowcolor[rgb]{0.867,0.922,0.969}
        \textbf{QGLAS}
            & \textbf{25.96} & \textbf{-0.93}
            & \textbf{20.40} & \textbf{-0.63}
            & \textbf{50.47} & \textbf{+2.27}
            & \textbf{32.28} & 2845
            & \textbf{102.0\%} & 32.4\% \\

        \midrule

        \multicolumn{11}{l}{\textbf{\textit{GLM-4.7-Flash}}} \\

        Base
            & 53.00 & --
            & 28.33 & --
            & 47.53 & --
            & 42.95 & 6571
            & 0.0\% & 8.7\% \\

        NoBonus
            & 40.11 & 0.00
            & 60.23 & 0.00
            & 80.57 & 0.00
            & 60.30 & 7199
            & 100.0\% & 0.0\% \\

        \hdashline
        \noalign{\vskip 1mm}

        GR$^3$
            & 36.56 & -3.55
            & 50.03 & -10.20
            & 78.83 & -1.74
            & 55.14 & 4780
            & 70.2\% & 33.6\% \\

        GRLC
            & 37.33 & -2.78
            & 49.17 & -11.06
            & 77.90 & -2.67
            & 54.80 & 4935
            & 68.3\% & 31.5\% \\

        \rowcolor[rgb]{0.867,0.922,0.969}
        \textbf{QGLAS}
            & \textbf{39.00} & \textbf{-1.11}
            & \textbf{59.30} & \textbf{-0.93}
            & \textbf{81.80} & \textbf{+1.23}
            & \textbf{60.03} & 4904
            & \textbf{98.4\%} & 31.9\% \\

        \bottomrule
    \end{tabular}%
}
\end{table*}

\paragraph{Matched-compression comparison.}
Table~\ref{tab:main_results} compares all three methods at approximately
matched compression. QGLAS retains 102.0\% and 98.4\% of the
macro-average quality gains of quality-only RL on Qwen3-4B and
GLM-4.7-Flash, respectively, compared with 75.5\% and 70.2\%
for GR$^3$, and 74.9\% and 68.3\% for GRLC. While QGR measures
aggregate rather than task-wise retention, QGLAS achieves the highest
mean quality score among length-controlled methods on every evaluated
benchmark for both policy models. Together with the benchmark-level
trade-offs in Appendix~\ref{app:benchmark_pareto}, this indicates that
the aggregate advantage is not driven solely by a single task.
Most of the observed IFBench degradation occurs already under
quality-only RL, before QGLAS introduces length control;
Appendix~\ref{app:benchmark_pareto} examines this behavior further.

\begin{figure*}[t]
    \centering
    \begin{subfigure}[t]{0.48\textwidth}
        \centering
        \includegraphics[width=\linewidth]{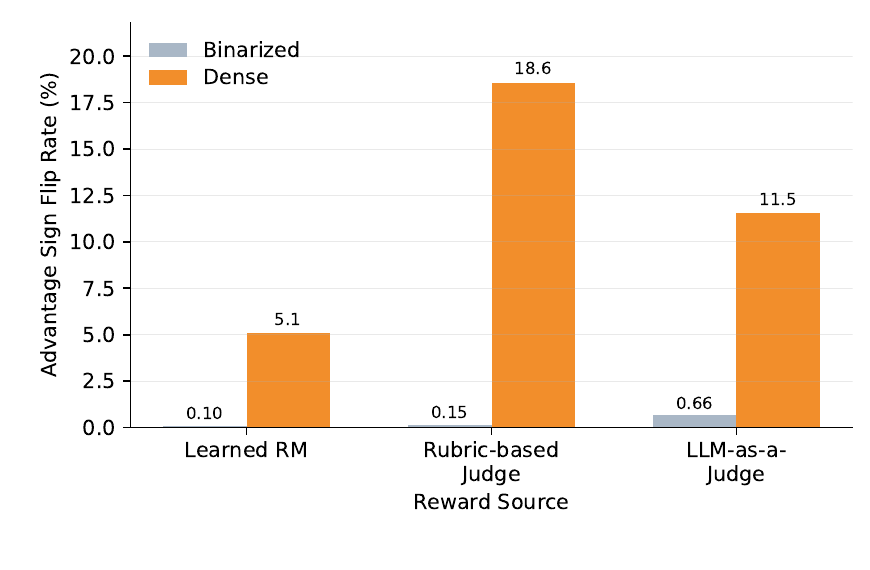}
        \caption{Sensitivity to reward granularity.}
        \label{fig:flip_granularity}
    \end{subfigure}
    \hfill
    \begin{subfigure}[t]{0.48\textwidth}
        \centering
        \includegraphics[width=\linewidth]{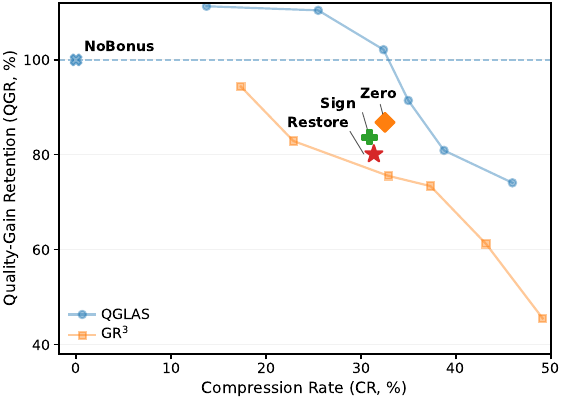}
        \caption{Targeted correction during training.}
        \label{fig:flip_correction}
    \end{subfigure}
    \caption{
        Advantage sign reversals: prevalence and downstream impact.
        \textbf{Left}: GR$^3$ flip rates under dense and binarized
        feedback on identical rollouts within each reward source.
        \textbf{Right}: Three corrections to GR$^3$ with $\alpha=0.3$
        on Qwen3-4B, shown against the GR$^3$ and QGLAS sweeps.
        Higher QGR and CR are preferred.
    }
    \label{fig:flip_analysis}
\end{figure*}

\subsection{Advantage Sign Reversals: Prevalence and Impact}
\label{sec:sign_reversal_impact}

Focusing on GR$^3$ as our closest baseline, we first diagnose sensitivity
to reward granularity on fixed rollouts, then intervene during training
to test whether correcting sign reversals improves its quality--length
trade-off.

\subsubsection{Sensitivity to Reward Granularity}
\label{sec:reward_granularity}

We use stored rollouts from steps 100--900 under three reward sources:
GR$^3$ training with $\alpha=0.3$ for Learned RM, and the corresponding
NoBonus runs for Rubric-based Judge and LLM-as-a-Judge.
Within each source, we hold responses, lengths, and rollout groups fixed
and compare the original dense rewards with four binarizations:
$\mathbf{1}[A_i^q>0]$ and indicators for the top $25\%$, $50\%$, or
$75\%$ of responses by quality score. We apply GR$^3$ with
$\alpha=0.3$ to each representation and average binarized results
over the four constructions.

A strict sign reversal satisfies
$A_i^{\mathrm{before}}A_i^{\mathrm{after}}<0$; transitions involving
zero are excluded. Each shaped advantage is compared with its own
pre-shaping counterpart under the same reward representation.
This isolates sensitivity to reward representation on fixed rollouts;
it does not reproduce RLVR training.

Figure~\ref{fig:flip_granularity} shows dense-feedback flip rates of
$5.08\%$, $18.56\%$, and $11.54\%$ for Learned RM, Rubric-based Judge,
and LLM-as-a-Judge, respectively, compared with $0.10\%$, $0.15\%$,
and $0.66\%$ after binarization. The macro average rises from
$0.30\%$ to $11.73\%$, a nearly $40\times$ difference. Thus,
the same reward-level shaping rule can reverse quality-induced
reinforcement polarity substantially more often under dense feedback.

\subsubsection{Targeted Correction of Sign Reversals}
\label{sec:flip_correction}

To test whether these reversals matter downstream, we train three
variants of GR$^3$ with $\alpha=0.3$ on Qwen3-4B under the main Learned
RM setting. Let $A_i^{\mathrm{GR}}$ denote the GR$^3$-shaped advantage.
Only when $A_i^q A_i^{\mathrm{GR}}<0$, we replace it by
$0$ (\textsc{Zero}),
$\operatorname{sign}(A_i^q)|A_i^{\mathrm{GR}}|$ (\textsc{Sign}), or
$A_i^q$ (\textsc{Restore}); all other advantages remain unchanged.
\textsc{Zero} suppresses the conflicting update, whereas
\textsc{Sign} and \textsc{Restore} recover its quality-induced polarity
using the shaped and original magnitudes, respectively.

Figure~\ref{fig:flip_correction} shows $(\mathrm{QGR},\mathrm{CR})$
values of $(86.82\%,32.53\%)$, $(83.56\%,30.92\%)$, and
$(80.02\%, 31.36\%)$ for \textsc{Zero}, \textsc{Sign}, and
\textsc{Restore}, respectively, compared with approximately
$(75.50\%,32.9\%)$ for unmodified GR$^3$.
All three corrections improve quality retention at similar compression
levels and lie above the interpolated GR$^3$ sweep.
\textsc{Zero} achieves both the highest QGR and the greatest compression
among the three corrections, recovering substantial quality at nearly
unchanged compression relative to unmodified GR$^3$.
These interventions support reversal-related interference as a
contributor to the quality cost of length shaping in this setting.
Nevertheless, all three remain below the QGLAS trade-off curve,
indicating that correcting sign reversals alone does not recover its
full advantage and motivating the following analysis of structural
constraints and adaptive strength.

\subsection{Preserving Direction and Adapting Magnitude}
\label{sec:ablation}

We separate QGLAS's structural constraints from its adaptive scaling,
examining both their overall effects and individual components.
Table~\ref{tab:ablation} reports downstream results under the main
Learned RM setting. For structural ablations, we vary only the scalar
shaping strength to approximately match full QGLAS's compression rate;
adaptive ablations use fixed statistics and achieve slightly weaker
compression. Appendix~\ref{app:full_ablation} provides calibration
details and unmatched results. The complementary offline flip rates
below are macro-averaged across three reward sources.

\begin{table*}[t]
\centering
\caption{
Ablation of QGLAS design choices on Qwen3-4B at approximately
matched compression. Each ablation modifies full QGLAS
independently.
IFB.\ denotes IFBench, Hard denotes Hard Prompts, and
Creative denotes Creative Writing.
Best quality results among length-controlled variants
are shown in bold.
}
\label{tab:ablation}

\resizebox{\textwidth}{!}{%
    \begin{tabular}{lccccccc}
        \toprule
        \textbf{Method}
        & \textbf{IFB.}$\uparrow$
        & \textbf{Hard}$\uparrow$
        & \textbf{Creative}$\uparrow$
        & \textbf{Avg. Score}$\uparrow$
        & \textbf{Avg. \#Tok.}$\downarrow$
        & \textbf{QGR}$\uparrow$
        & \textbf{CR}$\uparrow$ \\
        \midrule

        NoBonus
        & 26.89 & 21.03 & 48.20
        & 32.04 & 4207 & 100.0\% & 0.0\% \\

        \textbf{QGLAS}
        & \textbf{25.96}
        & \textbf{20.40}
        & \textbf{50.47}
        & \textbf{32.28}
        & 2845
        & \textbf{102.0\%}
        & 32.4\% \\

        \midrule
        \multicolumn{8}{l}{
            \textit{A. Structural constraints
            (adaptive scaling retained)}
        } \\
        \addlinespace[2pt]

        w/o All Structural Constraints
        & 23.33 & 16.13 & 42.77
        & 27.41 & 2895 & 60.0\% & 31.2\% \\

        \quad w/o Advantage-Level Shaping
        & 25.33 & 19.93 & 46.67
        & 30.64 & 2840 & 87.9\% & 32.5\% \\

        \quad w/o Positive-Only Gating
        & 24.89 & 17.83 & 44.57
        & 29.10 & 2799 & 74.6\% & 33.5\% \\

        \quad w/o One-Sided Shaping
        & 24.11 & 18.43 & 47.80
        & 30.11 & 2782 & 83.4\% & 33.9\% \\

        \midrule
        \multicolumn{8}{l}{
            \textit{B. Adaptive scaling
            (structural constraints retained)}
        } \\
        \addlinespace[2pt]

        Fixed Overall Strength ($\lambda=0.1833$)
        & 25.33 & 17.93 & 46.60
        & 29.95 & 2937 & 82.0\% & 30.2\% \\

        \quad Fixed $s_+=0.3234$
        & 25.56 & 18.40 & 48.30
        & 30.75 & 2955 & 88.9\% & 29.8\% \\

        \quad Fixed $s_{\mathrm{all}}=0.4672$
        & 25.11 & 18.03 & 47.67
        & 30.27 & 2960 & 84.7\% & 29.6\% \\

        \bottomrule
    \end{tabular}%
}
\end{table*}

\subsubsection{Structural Constraints for Quality-Aligned Shaping}
\label{sec:structural_ablation}

\paragraph{Are structural constraints necessary?}
Removing all three constraints applies two-sided length shaping at
the reward level to all responses while retaining the adaptive
strength rule. QGR falls from $102.0\%$ to $60.0\%$ at comparable
compression ($32.4\%$ vs.\ $31.2\%$), showing that adaptive strength
alone does not recover the full method's quality retention.
We next remove each constraint individually.

\paragraph{Isolating the length signal.}
Moving shaping from the advantage level to the reward level reduces
QGR to $87.9\%$ at $32.5\%$ compression. The corresponding offline
diagnostic reports a $0.99\%$ sign-reversal rate.
As discussed in Section~\ref{sec:quality_direction}, reward-level
shaping changes the group statistics used to form advantages and
can therefore affect responses without a direct length bonus.
Advantage-level shaping avoids this indirect interference.

\paragraph{Restricting the scope of shaping.}
Removing positive-only gating reduces QGR to $74.6\%$ at $33.5\%$
compression; allowing two-sided shaping reduces it to $83.4\%$ at
$33.9\%$ compression. Their offline flip rates are $3.80\%$ and
$2.71\%$, respectively. Without positive-only gating, brevity bonuses
can counteract negative quality advantages; two-sided shaping can
instead turn positive advantages negative for longer responses.
These results support restricting conciseness bonuses to
quality-favored responses while leaving longer ones unpenalized.

\subsubsection{Quality-Adaptive Scaling Beyond Sign Preservation}
\label{sec:adaptive_scaling}
\label{sec:positive_pair_reversal}

\paragraph{Does adapting the magnitude matter?}
All adaptive-scaling ablations retain the structural constraints
and preserve every advantage sign.
Replacing $\lambda_g$ by its mean over stored full-QGLAS rollout
groups, $\lambda_{\mathrm{fixed}}=0.1833$, reduces QGR from
$102.0\%$ to $82.0\%$ despite weaker compression
($30.2\%$ vs.\ $32.4\%$). Thus, sign preservation alone does not
ensure strong quality retention; adapting bonus magnitudes across
groups provides an additional benefit.

\paragraph{Which reward statistics matter?}
We separately fix $s_+$ or $s_{\mathrm{all}}$, retaining the other
statistic's group-specific value.
Fixing $s_+$ yields $88.9\%$ QGR at $29.8\%$ compression.
Fixing $s_{\mathrm{all}}$ in both the outer scale factor and the
computation of $\beta_g$ yields $84.7\%$ QGR at $29.6\%$ compression.
Both variants retain more quality than fixed overall strength, but
remain below full QGLAS despite weaker compression.
These results support using both the overall reward spread and
the quality separation among positive-advantage responses to
calibrate shaping strength.

\paragraph{Where does the reinforcement ordering change?}
QGLAS preserves advantage signs but allows conciseness to change
the ordering within the positive-advantage set.
For pairs with $A_i^q>A_j^q>0$, we count a ranking reversal when
$\widetilde{A}_i<\widetilde{A}_j$ and divide eligible pairs into five
equal-sized bins by their pre-shaping advantage gap,
$\Delta_{ij}^q=A_i^q-A_j^q$.
Table~\ref{tab:positive_pair_reversal} shows that reordering is
concentrated among small-gap pairs; fewer than $1\%$ of pairs in
the largest-gap bin are reordered under each reward source.
This characterizes how full QGLAS reallocates reinforcement
primarily among weakly separated responses, complementing the
downstream evidence for adaptive scaling above.

\begin{table*}[t]
\centering
\caption{
Robustness across training quality rewards on Qwen3-4B.
IFB.\ denotes IFBench, Hard denotes the Hard Prompts subset of
Arena-Hard-v2, and Creative denotes its Creative Writing subset.
}
\label{tab:reward_robustness}

\resizebox{\textwidth}{!}{%
    \begin{tabular}{llccccccc}
        \toprule
        \textbf{Training Reward}
        & \textbf{Method}
        & \textbf{IFB.}$\uparrow$
        & \textbf{Hard}$\uparrow$
        & \textbf{Creative}$\uparrow$
        & \textbf{Avg. Score}$\uparrow$
        & \textbf{Avg. \#Tok.}$\downarrow$
        & \textbf{QGR}$\uparrow$
        & \textbf{CR}$\uparrow$ \\
        \midrule

        \multirow{2}{*}{Learned RM}
        & NoBonus
        & \textbf{26.89} & \textbf{21.03} & 48.20
        & 32.04 & 4207
        & -- & -- \\
        & \textbf{QGLAS}
        & 25.96 & 20.40 & \textbf{50.47}
        & \textbf{32.28} & \textbf{2845}
        & \textbf{102.0\%} & \textbf{32.4\%} \\
        \midrule

        \multirow{2}{*}{Rubric-based Judge}
        & NoBonus
        & 33.11 & \textbf{19.07} & \textbf{28.87}
        & \textbf{27.02} & 4125
        & -- & -- \\
        & \textbf{QGLAS}
        & \textbf{33.44} & 18.93 & 28.57
        & 26.98 & \textbf{2960}
        & \textbf{99.4\%} & \textbf{28.2\%} \\
        \midrule

        \multirow{2}{*}{LLM-as-a-Judge}
        & NoBonus
        & \textbf{32.88} & \textbf{17.13} & 23.50
        & \textbf{24.50} & 4248
        & -- & -- \\
        & \textbf{QGLAS}
        & 32.67 & 16.87 & \textbf{23.90}
        & 24.48 & \textbf{3016}
        & \textbf{99.4\%} & \textbf{29.0\%} \\
        \bottomrule
    \end{tabular}%
}
\end{table*}

\subsection{Generalization Across Reward Sources}
\label{sec:reward_robustness}

Finally, we test whether QGLAS's benefit depends on the training reward.
We replace the Learned RM with Rubric-based Judge and LLM-as-a-Judge
feedback, keeping the remaining training setup and QGLAS
hyperparameters unchanged.
Table~\ref{tab:reward_robustness} shows that QGLAS reduces average
response length by $28.2$--$32.4\%$ across the three reward sources
while retaining $99.4$--$102.0\%$ of the corresponding quality-only
RL gains. These results support robustness across a learned reward model and two forms of LLM-based feedback without reward-specific retuning.

\section{Conclusion}

In this work, we study length control in open-ended reinforcement learning, where dense quality feedback and the strong coupling between quality and response length make existing RLVR-style solutions difficult to apply directly. We argue that length control should respect the reinforcement direction established by the original quality feedback, rather than allowing length to reverse it. Based on this view, we propose three principles for quality-aligned length control: isolating the length signal, restricting where it can act, and adapting its strength to the local quality structure. We instantiate these principles in QGLAS, which consistently achieves a better quality--length trade-off across evaluated models, benchmarks, and reward sources. At around 30\% compression, QGLAS retains nearly all of the macro-average quality improvement achieved by the corresponding quality-only RL policy over the base model, suggesting that conciseness is best treated as a secondary preference under quality rather than as a competing objective.

\bibliography{iclr2027_conference}
\bibliographystyle{iclr2027_conference}

\appendix
\clearpage

\section{Experimental Details}
\label{app:experimental_details}

\subsection{Training Configuration}
\label{app:training_details}

Unless otherwise specified, NoBonus, GR$^3$, GRLC, and QGLAS use the same
training configuration within each policy-model comparison. The main training
hyperparameters are summarized in Table~\ref{tab:training_details}.

\begin{table}[t]
\centering
\small
\caption{Training configuration for the main experiments.}
\label{tab:training_details}
\begin{tabular}{lc}
\toprule
Hyperparameter & Value \\
\midrule
RL algorithm & GSPO \\
Optimizer & Adam \\
Learning rate & $1\times10^{-6}$ \\
Learning-rate schedule & Constant \\
Adam $(\beta_1,\beta_2)$ & $(0.9, 0.98)$ \\
Weight decay & 0.1 \\
Rollout batch size & 32 \\
Mini-batch size & 32 \\
Global batch size & 512 \\
Responses per prompt & 16 \\
Rollout temperature & 1.0 \\
Maximum prompt length & 2,048 \\
Maximum response length & 8,192 \\
KL coefficient & 0.001 \\
Entropy coefficient & 0 \\
Policy clip range & $3\times10^{-4}$ / $4\times10^{-4}$ \\
Default seed (unless otherwise specified) & 42 \\
\bottomrule
\end{tabular}
\end{table}

We disable group-wise standard-deviation normalization of the quality
advantage in our GSPO implementation
\citep{kimi2.5,dr.grpo,li2025darling}. Accordingly, QGLAS uses the
non-standardized form of the group-level shaping scale described in
Section~\ref{sec:adaptive_bonus}.

For the main matched-compression comparison on Qwen3-4B, we train NoBonus,
GR$^3$ with $\alpha=0.3$, GRLC with $(\lambda,\beta)=(0.2,0.2)$, and
QGLAS with $(\beta_{\min},\beta_{\max})=(0.3,0.6)$ using three seeds
(42, 43, and 44). The same seeds are used across all four methods, and each
trained policy is evaluated three times. The Qwen3-4B matched-compression
results reported in the main text are averaged across these three training
seeds. For GLM-4.7-Flash, each configuration is trained once because of its
substantially higher training cost, and each trained policy is evaluated
three times.

\paragraph{Compute resources.}
Qwen3-4B policies are trained on 8 NVIDIA H800 GPUs, whereas
GLM-4.7-Flash policies are trained on 16 H800 GPUs. In experiments using
the learned reward model, Skywork-Reward-V2-Llama-3.1-8B is served on an
additional 2 H800 GPUs. The resulting compute usage is approximately
160--240 H800 GPU-hours per Qwen3-4B training run and 576--864 H800
GPU-hours per GLM-4.7-Flash training run.

The Rubric-based Judge and LLM-as-a-Judge reward experiments are conducted
only with Qwen3-4B, with Qwen3.5-35B-A3B served on an additional 8 H800
GPUs. Each such training run uses approximately 768 H800 GPU-hours.
Compute usage varies with rollout length, with configurations inducing
stronger compression generally requiring fewer GPU-hours.

\paragraph{Alternative reward sources.}
For Rubric-based Judge and LLM-as-a-Judge, we follow the reward
construction procedures of \citet{weng2026prompt}, using normalized
weighted aggregation of rubric-level judgments and global scores
normalized to $[0,1]$, respectively.
Both use Qwen3.5-35B-A3B as the judge.

\subsection{Length-Control Configuration}
\label{app:length_control_details}

\paragraph{QGLAS.}
Unless otherwise specified, we use the same QGLAS-specific hyperparameters
across experiments. For the response-level length coefficient in
\eqref{eq:length_gain}, we set the relative-shortening clipping threshold to
$c=0.5$ and the numerical stability constant to $\epsilon=10^{-8}$. The
overall reward spread $s_{\mathrm{all}}$ is computed from the 25th and
100th percentiles of the within-group quality rewards.

The main QGLAS configuration uses
$(\beta_{\min},\beta_{\max})=(0.3,0.6)$. In the control-strength sweeps,
we set $\beta_{\max}=2\beta_{\min}$ and vary only $\beta_{\min}$.
All other QGLAS-specific hyperparameters are held fixed across policy
models and reward sources.

\paragraph{Normalization and clipping behavior.}
We fix the relative-shortening scale to $c=0.5$ in
\eqref{eq:length_gain}. Besides determining the saturation threshold,
$c$ also normalizes the response-level shortening coefficient. Let
$d_i=(L_{\mathrm{ref}}-L_i)/(L_{\mathrm{ref}}+\epsilon)$. In the
unsaturated regime, $0<d_i<c$, we have $h_i=d_i/c$, so $h_i$ represents
the fraction of the shortening scale $c$ attained by response $i$ and
reaches its maximum value of $1$ when $d_i\geq c$. For example, with
$c=0.5$, a $10\%$ relative shortening corresponds to $h_i=0.2$.
Since the group-level shaping scale $\lambda_g$ is proportional to
$\beta_g$, the effective shaping strength in this linear regime depends
on $\beta_g/c$; we therefore keep $c$ fixed and vary $\beta_g$ to control
the overall strength of length shaping.

We further examine how often the saturation threshold is reached in
practice. Using stored rollout groups from steps 100--900 of the
Qwen3-4B QGLAS training runs with
$(\beta_{\min},\beta_{\max})=(0.3,0.6)$ under each of the three reward
sources, we consider only responses receiving a non-zero length bonus
($h_i>0$) and measure the fraction whose pre-clipping relative-shortening
term exceeds $c$.

As shown in Table~\ref{tab:clip_activation}, the threshold is reached by
only 0.719\%, 0.790\%, and 1.286\% of bonus-receiving responses under the
Learned RM, Rubric-based Judge, and LLM-as-a-Judge settings, respectively.
Pooled across the three reward sources, only 3,043 of 331,491 such
responses (0.918\%) reach the threshold. Thus, more than 99\% of shaped
responses operate in the linear, unsaturated regime, indicating that the
observed behavior of QGLAS is not driven by frequent saturation at the
chosen threshold.

\begin{table}[t]
\centering
\small
\caption{
Activation frequency of the clipping threshold $c=0.5$ in
\eqref{eq:length_gain}. Statistics are computed from stored rollout groups
from steps 100--900 of QGLAS
$(\beta_{\min},\beta_{\max})=(0.3,0.6)$ training. The denominator includes
only responses with $h_i>0$; ``Clipped'' denotes responses whose
pre-clipping relative-shortening term exceeds $c$.
}
\label{tab:clip_activation}
\begin{tabular}{lrrr}
\toprule
Reward source & \#$h_i>0$ & Clipped & Clip rate \\
\midrule
Learned RM         & 111,418 &   801 & 0.719\% \\
Rubric-based Judge & 118,504 &   936 & 0.790\% \\
LLM-as-a-Judge     & 101,569 & 1,306 & 1.286\% \\
\midrule
Pooled             & 331,491 & 3,043 & 0.918\% \\
\bottomrule
\end{tabular}
\end{table}

\paragraph{GR$^3$.}
Following the sigmoid reward transformation used in the RLHF setting of
GR$^3$, we apply PAR-style shaping to the raw reward-model scores:
\begin{equation}
r_i
=
\sigma\!\left(
\frac{r_i^{\mathrm{raw}}-r_{\mathrm{ref}}^{\mathrm{raw}}}{\tau}
\right).
\end{equation}
We use the within-group median raw reward as
$r_{\mathrm{ref}}^{\mathrm{raw}}$ and fix the temperature at $\tau=2$
to mitigate sigmoid saturation.
The resulting rewards are non-negative, and the same transformation
is applied to all methods in the corresponding comparisons.

GR$^3$ applies multiplicative length-dependent rescaling to the quality
reward:
\begin{equation}
\hat{r}_i
=
\frac{r_i}{1+\alpha L_i/\bar{L}},
\label{eq:gr3_appendix}
\end{equation}
where $L_i$ is the response length, $\bar{L}$ is the mean response length
within the rollout group, and $\alpha$ controls the strength of length
regularization.

GR$^3$ selects $\alpha$ using an advantage-preservation calibration
criterion. For a representative maximum-reward response with group-average
length, the shaped reward is required to satisfy
\begin{equation}
\frac{r_{\max}}{1+\alpha}
\geq
\mu_{\hat{r}},
\label{eq:gr3_calibration}
\end{equation}
where $\mu_{\hat{r}}$ denotes the group mean of the shaped rewards.
Following the GR$^3$ calibration procedure, $\alpha$ is chosen as large
as possible while maintaining a sufficiently high constraint-satisfaction
rate across rollout groups. All coefficients in our GR$^3$ sweep, including
$\alpha=0.6$, achieve a constraint-satisfaction rate of at least 99.9\%.
The matched-compression comparison in Table~\ref{tab:main_results} uses
$\alpha=0.3$.

\paragraph{GRLC.}
GRLC~\citep{nvidia2025grlc} performs reward-level length shaping separately
for the reasoning and final-answer components of each response. Let
$\ell_i^{(\mathrm{think})}$ and $\ell_i^{(\mathrm{answer})}$ denote the
corresponding token lengths of response $i$. For each component
$c\in\{\mathrm{think},\mathrm{answer}\}$, GRLC first computes the
group-relative shortness weight
\begin{equation}
w_i^{(c)}
=
1-
\frac{
\ell_i^{(c)}-\ell_{\min}^{(c)}
}{
\ell_{\max}^{(c)}-\ell_{\min}^{(c)}
},
\label{eq:grlc_shortness}
\end{equation}
where
$\ell_{\min}^{(c)}=\min_j\ell_j^{(c)}$ and
$\ell_{\max}^{(c)}=\max_j\ell_j^{(c)}$.
The weights are then centered within each rollout group:
\begin{equation}
\widetilde{w}_i^{(c)}
=
w_i^{(c)}
-
\frac{1}{N}\sum_{j=1}^{N}w_j^{(c)},
\label{eq:grlc_centering}
\end{equation}
where $N$ is the rollout-group size. If all responses have the same length
for a component, we set the corresponding centered weights to zero.

The resulting length-adjusted reward is
\begin{equation}
r_i^{\mathrm{len}}
=
r_i
+
\lambda^{(\mathrm{think})}
\widetilde{w}_i^{(\mathrm{think})}
+
\lambda^{(\mathrm{answer})}
\widetilde{w}_i^{(\mathrm{answer})}.
\label{eq:grlc_length_reward}
\end{equation}

GRLC additionally applies quality-gated conciseness bonuses to the responses
with the shortest reasoning trace and shortest final answer. Let
\begin{equation}
k_{\mathrm{think}}
=
\arg\min_j \ell_j^{(\mathrm{think})},
\qquad
k_{\mathrm{answer}}
=
\arg\min_j \ell_j^{(\mathrm{answer})},
\end{equation}
and let $\tau_p$ denote the $p$-th percentile of the within-group base
quality rewards. The final shaped reward is
\begin{align}
\widehat{r}_i
={}&
r_i^{\mathrm{len}}
+
\beta^{(\mathrm{think})}
\mathbb{I}[i=k_{\mathrm{think}}]
\mathbb{I}[r_i\geq\tau_p]
\nonumber\\
&+
\beta^{(\mathrm{answer})}
\mathbb{I}[i=k_{\mathrm{answer}}]
\mathbb{I}[r_i\geq\tau_p].
\label{eq:grlc_appendix}
\end{align}
A response that is shortest in both components may therefore receive both
bonuses.

The original GRLC configuration sets
$\lambda^{(\mathrm{think})}=\lambda^{(\mathrm{answer})}=0.5$ and
$\beta^{(\mathrm{think})}=\beta^{(\mathrm{answer})}=0.5$, with
$p=80$~\citep{nvidia2025grlc}. In our experiments, we keep $p=80$ fixed
and vary only the overall length-control strength. Specifically, we tie the
coefficients across the reasoning and final-answer components:
\[
\lambda^{(\mathrm{think})}
=
\lambda^{(\mathrm{answer})}
=
\lambda,
\qquad
\beta^{(\mathrm{think})}
=
\beta^{(\mathrm{answer})}
=
\beta,
\]
and jointly sweep
\[
(\lambda,\beta)
\in
\{
(0.1,0.1),
(0.15,0.15),
(0.2,0.2),
(0.25,0.25),
(0.3,0.3)
\}.
\]
For the matched-compression comparison in
Table~\ref{tab:main_results}, we use $(\lambda,\beta)=(0.2,0.2)$,
which gives the compression rate closest to the main QGLAS operating point
among the evaluated GRLC configurations.

Following the original formulation, reasoning and final-answer lengths are
shaped separately rather than combined into total response length. We split
each response at the first \texttt{</think>} marker. If the marker is absent,
the entire generated response is treated as the reasoning component and the
final-answer length is set to zero.

\subsection{Evaluation Configuration}
\label{app:evaluation_details}

Each trained policy is evaluated three times under the benchmark-specific
evaluation protocol. We report the resulting average quality score and
average response length.

\paragraph{IFBench.}
For IFBench, responses are generated with temperature $0.6$, top-$p=0.95$,
top-$k=20$, and a maximum response length of 16384 tokens. We follow the
standard IFBench evaluation procedure and report the \emph{prompt-level
strict} metric throughout the paper.

\paragraph{Arena-Hard-v2.}
For Arena-Hard-v2, responses are generated with temperature $0.6$ and a
maximum response length of 32,000 tokens. We do not apply top-$p$, top-$k$,
or min-$p$ truncation. We follow the official Arena-Hard-v2 evaluation
procedure and use GPT-4.1 for judge-based evaluation. The judge uses
deterministic decoding with temperature $0$ and a maximum output length of
16,000 tokens.

\section{Additional Results}
\label{app:additional_results}

\subsection{Repeated Qwen3-4B Runs}
\label{app:qwen_repeated}

We evaluate the stability of the main Qwen3-4B matched-compression
comparison across three training seeds (42, 43, and 44). The same seeds
are used for NoBonus, GR$^3$ with $\alpha=0.3$, GRLC with
$(\lambda,\beta)=(0.2,0.2)$, and QGLAS with
$(\beta_{\min},\beta_{\max})=(0.3,0.6)$. All other training settings are
held fixed. Each trained policy is evaluated three times, and the
evaluation results are first averaged within each training run.
Table~\ref{tab:qwen_repeated} reports the individual training runs and
the mean $\pm$ sample standard deviation across seeds.

\begin{table*}[t]
\centering
\caption{
Repeated Qwen3-4B results across three training seeds. Mean rows report
mean $\pm$ sample standard deviation across seeds. Each trained policy is
evaluated three times, with evaluation results averaged within each
training run.
}
\label{tab:qwen_repeated}

\resizebox{\textwidth}{!}{%
\begin{tabular}{lcccccccc}
\toprule
& \multicolumn{2}{c}{IFBench}
& \multicolumn{2}{c}{Hard Prompts}
& \multicolumn{2}{c}{Creative Writing}
& \multicolumn{2}{c}{Macro Average} \\
\cmidrule(lr){2-3}
\cmidrule(lr){4-5}
\cmidrule(lr){6-7}
\cmidrule(lr){8-9}
Method
& Score $\uparrow$
& \#Tok. $\downarrow$
& Score $\uparrow$
& \#Tok. $\downarrow$
& Score $\uparrow$
& \#Tok. $\downarrow$
& Score $\uparrow$
& \#Tok. $\downarrow$ \\
\midrule

\textbf{NoBonus}
& $26.89 \pm 0.29$
& $2834 \pm 65$
& $21.03 \pm 0.32$
& $7149 \pm 123$
& $48.20 \pm 0.79$
& $2638 \pm 61$
& $32.04 \pm 0.47$
& $4207 \pm 83$ \\

\quad Run 1 (seed 42)
& 26.78 & 2816
& 20.90 & 7105
& 47.90 & 2620
& 31.86 & 4180 \\

\quad Run 2 (seed 43)
& 27.22 & 2780
& 21.40 & 7054
& 49.10 & 2588
& 32.57 & 4141 \\

\quad Run 3 (seed 44)
& 26.67 & 2906
& 20.80 & 7288
& 47.60 & 2706
& 31.69 & 4300 \\

\midrule

\textbf{GR$^3$}
& $25.07 \pm 0.34$
& $1403 \pm 48$
& $17.07 \pm 0.38$
& $5296 \pm 116$
& $45.47 \pm 0.83$
& $1774 \pm 60$
& $29.20 \pm 0.51$
& $2824 \pm 74$ \\

\quad Run 1 (seed 42)
& 25.00 & 1390
& 16.90 & 5258
& 45.20 & 1758
& 29.03 & 2802 \\

\quad Run 2 (seed 43)
& 24.78 & 1363
& 16.80 & 5204
& 44.80 & 1724
& 28.79 & 2764 \\

\quad Run 3 (seed 44)
& 25.44 & 1456
& 17.50 & 5426
& 46.40 & 1840
& 29.78 & 2907 \\

\midrule

\textbf{GRLC}
& $24.78 \pm 0.29$
& $1517 \pm 62$
& $20.13 \pm 0.42$
& $5447 \pm 129$
& $42.50 \pm 0.89$
& $1668 \pm 61$
& $29.14 \pm 0.53$
& $2877 \pm 84$ \\

\quad Run 1 (seed 42)
& 24.67 & 1500
& 20.00 & 5408
& 42.20 & 1650
& 28.96 & 2853 \\

\quad Run 2 (seed 43)
& 24.56 & 1465
& 19.80 & 5342
& 41.80 & 1618
& 28.72 & 2808 \\

\quad Run 3 (seed 44)
& 25.11 & 1586
& 20.60 & 5591
& 43.50 & 1736
& 29.74 & 2971 \\

\midrule

\textbf{QGLAS}
& $25.96 \pm 0.42$
& $1432 \pm 62$
& $20.40 \pm 0.36$
& $5379 \pm 120$
& $50.47 \pm 0.93$
& $1723 \pm 56$
& $32.28 \pm 0.57$
& $2845 \pm 79$ \\

\quad Run 1 (seed 42)
& 25.78 & 1415
& 20.30 & 5342
& 50.20 & 1705
& 32.09 & 2821 \\

\quad Run 2 (seed 43)
& 26.44 & 1380
& 20.80 & 5282
& 51.50 & 1678
& 32.91 & 2780 \\

\quad Run 3 (seed 44)
& 25.67 & 1501
& 20.10 & 5513
& 49.70 & 1786
& 31.82 & 2933 \\

\bottomrule
\end{tabular}%
}
\end{table*}

The repeated runs reproduce the matched-compression advantage reported in
Table~\ref{tab:main_results}. GR$^3$, GRLC, and QGLAS achieve closely
matched macro-average response lengths of 2824, 2877, and 2845 tokens,
respectively, while QGLAS achieves a substantially higher macro-average
quality score of 32.28, compared with 29.20 for GR$^3$ and 29.14 for GRLC.

Importantly, this advantage is reproduced across all three training seeds.
At matched seeds, QGLAS exceeds GR$^3$ in macro-average quality by
$+3.06$, $+4.12$, and $+2.04$ points for seeds 42, 43, and 44,
respectively, and exceeds GRLC by $+3.13$, $+4.19$, and $+2.08$ points.
Thus, despite normal variation across training runs, QGLAS consistently
maintains a higher macro-average quality score at comparable response
lengths, indicating that the matched-compression advantage is not driven
by a particular training seed.

\subsection{Full Quality--Length Trade-off Results}
\label{app:full_pareto}

Table~\ref{tab:full_pareto} reports the complete Qwen3-4B
quality--length sweeps. For QGLAS, we vary $\beta_{\min}$ while setting
$\beta_{\max}=2\beta_{\min}$; for GR$^3$, we vary $\alpha$; and for
GRLC, we jointly vary $(\lambda,\beta)$. All other method-specific
hyperparameters are held fixed.

Each trained policy is evaluated three times. For the matched-compression
configurations, the reported values additionally average across the three
training seeds described in Section~\ref{app:qwen_repeated}. QGR and CR are
computed from the unrounded macro averages according to
\eqref{eq:relative_metrics}.

\begin{table*}[t]
\centering
\caption{
Full quality--length trade-off results on Qwen3-4B. Each benchmark reports
quality score and average response length. Parenthetical values denote
$(\beta_{\min},\beta_{\max})$ for QGLAS, $\alpha$ for GR$^3$, and
$(\lambda,\beta)$ for GRLC. QGR and CR are computed according to
\eqref{eq:relative_metrics}. Higher values are preferred for both metrics.
}
\label{tab:full_pareto}

\resizebox{\textwidth}{!}{%
\begin{tabular}{lcccccccccc}
\toprule
& \multicolumn{2}{c}{IFBench}
& \multicolumn{2}{c}{Hard Prompts}
& \multicolumn{2}{c}{Creative Writing}
& \multicolumn{2}{c}{Macro Average}
& \multicolumn{2}{c}{Relative Metrics} \\
\cmidrule(lr){2-3}
\cmidrule(lr){4-5}
\cmidrule(lr){6-7}
\cmidrule(lr){8-9}
\cmidrule(lr){10-11}
Method
& Score $\uparrow$
& \#Tok. $\downarrow$
& Score $\uparrow$
& \#Tok. $\downarrow$
& Score $\uparrow$
& \#Tok. $\downarrow$
& Score $\uparrow$
& \#Tok. $\downarrow$
& QGR $\uparrow$
& CR $\uparrow$ \\
\midrule

Base
& 29.22 & 2548
& 15.63 & 6823
& 16.53 & 2004
& 20.46 & 3792
& 0.0\% & 9.9\% \\

NoBonus
& 26.89 & 2834
& 21.03 & 7149
& 48.20 & 2638
& 32.04 & 4207
& 100.0\% & 0.0\% \\

\midrule

QGLAS $(0.1,0.2)$
& 26.56 & 2203
& 20.87 & 6382
& 52.63 & 2301
& 33.35 & 3629
& 111.3\% & 13.8\% \\

QGLAS $(0.2,0.4)$
& 26.11 & 1892
& 20.50 & 5596
& 53.10 & 1914
& 33.24 & 3134
& 110.3\% & 25.5\% \\

QGLAS $(0.3,0.6)$
& 25.96 & 1432
& 20.40 & 5379
& 50.47 & 1723
& 32.28 & 2845
& 102.0\% & 32.4\% \\

QGLAS $(0.4,0.8)$
& 24.89 & 1311
& 19.90 & 5262
& 48.30 & 1632
& 31.03 & 2735
& 91.3\% & 35.0\% \\

QGLAS $(0.5,1.0)$
& 24.56 & 1166
& 19.17 & 5100
& 45.77 & 1465
& 29.83 & 2577
& 80.9\% & 38.7\% \\

QGLAS $(0.6,1.2)$
& 24.22 & 1037
& 19.33 & 4503
& 43.57 & 1281
& 29.04 & 2274
& 74.1\% & 46.0\% \\

\midrule

GR$^3$ $(0.1)$
& 26.00 & 2120
& 20.07 & 6136
& 48.00 & 2176
& 31.36 & 3477
& 94.1\% & 17.3\% \\

GR$^3$ $(0.2)$
& 25.11 & 1815
& 20.10 & 5841
& 44.93 & 2076
& 30.05 & 3244
& 82.8\% & 22.9\% \\

GR$^3$ $(0.3)$
& 25.07 & 1403
& 17.07 & 5296
& 45.47 & 1774
& 29.20 & 2824
& 75.5\% & 32.9\% \\

GR$^3$ $(0.4)$
& 24.89 & 1347
& 16.80 & 5024
& 45.13 & 1540
& 28.94 & 2637
& 73.2\% & 37.3\% \\

GR$^3$ $(0.5)$
& 24.11 & 1129
& 16.73 & 4652
& 41.83 & 1391
& 27.56 & 2391
& 61.3\% & 43.2\% \\

GR$^3$ $(0.6)$
& 23.11 & 1017
& 15.27 & 4097
& 38.77 & 1306
& 25.72 & 2140
& 45.4\% & 49.1\% \\

\midrule

GRLC $(0.1,0.1)$
& 25.89 & 2105
& 20.90 & 6332
& 47.40 & 2133
& 31.40 & 3523
& 94.4\% & 16.3\% \\

GRLC $(0.15,0.15)$
& 25.67 & 1910
& 20.73 & 5856
& 43.97 & 1997
& 30.12 & 3254
& 83.5\% & 22.6\% \\

GRLC $(0.2,0.2)$
& 24.78 & 1517
& 20.13 & 5447
& 42.50 & 1668
& 29.14 & 2877
& 74.9\% & 31.6\% \\

GRLC $(0.25,0.25)$
& 24.44 & 1342
& 18.70 & 5132
& 41.57 & 1458
& 28.24 & 2644
& 67.2\% & 37.2\% \\

GRLC $(0.3,0.3)$
& 23.22 & 1126
& 16.93 & 4345
& 39.80 & 1236
& 26.65 & 2236
& 53.5\% & 46.9\% \\

\bottomrule
\end{tabular}%
}
\end{table*}

Across the overlapping compression range, QGLAS provides a consistently
stronger aggregate quality--length trade-off than GR$^3$ and GRLC. Near
32\% compression, QGLAS retains 102.0\% of the quality gain at 32.4\%
compression, compared with 75.5\% for GR$^3$ at 32.9\% compression and
74.9\% for GRLC at 31.6\%. The full sweep shows that this separation is
not specific to the matched-compression operating point.

\subsection{Additional Benchmark-Level Analysis}
\label{app:benchmark_pareto}

QGR is an aggregate retention metric and does not imply non-degradation
on every individual benchmark. In particular, the quality-only RL policy
(NoBonus) may itself improve some benchmarks while degrading others
relative to the base model.

IFBench illustrates this distinction. Under the Learned RM setting,
Qwen3-4B decreases from 29.22 for Base to 26.89 for NoBonus before any
length-control signal is introduced. QGLAS reaches 25.96 at 32.4\%
compression, only 0.93 points below NoBonus, while outperforming
GR$^3$ and GRLC at comparable compression. The pattern is similar on
GLM-4.7-Flash: IFBench decreases from 53.00 for Base to 40.11 for
NoBonus, while QGLAS further decreases it by 1.11 points to 39.00.
Thus, most of the observed IFBench degradation occurs under quality-only
RL rather than after introducing QGLAS.

The alternative reward sources provide a useful contrast. Under
Rubric-based Judge and LLM-as-a-Judge rewards, NoBonus achieves IFBench
scores of 33.11 and 32.88, respectively, both above the base-model score
of 29.22. QGLAS obtains 33.44 and 32.67 while substantially reducing
response length. This suggests that the benchmark-level behavior depends
primarily on the underlying quality objective rather than reflecting a
systematic degradation induced by QGLAS.

Finally, on Creative Writing, the two mildest QGLAS settings improve
quality over NoBonus while simultaneously reducing response length,
showing that the quality--length trade-off need not be strictly zero-sum
in every operating regime.

\subsection{Full Ablation Results and Compression Matching}
\label{app:full_ablation}

\begin{table*}[t]
\centering
\caption{
Full QGLAS ablation results on Qwen3-4B.
Structural ablations include both original and approximately
compression-matched configurations; parenthetical values in
``Matched'' rows denote the scalar shaping-strength setting.
Adaptive ablations use the fixed statistics shown in the
Setting column. Each ablation modifies full QGLAS independently.
Best quality results among length-controlled variants are bold.
}
\label{tab:full_ablation}

\resizebox{\textwidth}{!}{%
    \begin{tabular}{llrrrrrrrrrr}
        \toprule
        & &
        \multicolumn{2}{c}{IFBench} &
        \multicolumn{2}{c}{Hard Prompts} &
        \multicolumn{2}{c}{Creative Writing} &
        \multicolumn{2}{c}{Macro Average} &
        \multicolumn{2}{c}{Relative Metrics} \\
        \cmidrule(lr){3-4}
        \cmidrule(lr){5-6}
        \cmidrule(lr){7-8}
        \cmidrule(lr){9-10}
        \cmidrule(lr){11-12}

        Variant & Setting
        & Score$\uparrow$ & \#Tok.$\downarrow$
        & Score$\uparrow$ & \#Tok.$\downarrow$
        & Score$\uparrow$ & \#Tok.$\downarrow$
        & Score$\uparrow$ & \#Tok.$\downarrow$
        & QGR$\uparrow$ & CR$\uparrow$ \\
        \midrule

        Base
        & --
        & 29.22 & 2548
        & 15.63 & 6823
        & 16.53 & 2004
        & 20.46 & 3792
        & 0.0\% & 9.9\% \\

        NoBonus
        & --
        & 26.89 & 2834
        & 21.03 & 7149
        & 48.20 & 2638
        & 32.04 & 4207
        & 100.0\% & 0.0\% \\

        \textbf{QGLAS}
        & $(0.3,0.6)$
        & \textbf{25.96} & 1432
        & \textbf{20.40} & 5379
        & \textbf{50.47} & 1723
        & \textbf{32.28} & 2845
        & \textbf{102.0\%} & 32.4\% \\

        \midrule
        \multicolumn{12}{l}{
            \textit{A. Structural constraints
            (adaptive scaling retained)}
        } \\
        \addlinespace[2pt]

        \multirow{2}{*}{
            \shortstack[l]{w/o All Structural\\Constraints}
        }
        & Original
        & 23.00 & 734
        & 15.43 & 3498
        & 34.70 & 1032
        & 24.38 & 1755
        & 33.8\% & 58.3\% \\

        & Matched (0.35)
        & 23.33 & 1576
        & 16.13 & 5311
        & 42.77 & 1799
        & 27.41 & 2895
        & 60.0\% & 31.2\% \\

        \addlinespace[3pt]

        \multirow{2}{*}{
            \shortstack[l]{
                \quad w/o Advantage-Level\\
                \quad Shaping
            }
        }
        & Original
        & 24.89 & 1614
        & 20.17 & 5922
        & 48.87 & 1926
        & 31.31 & 3154
        & 93.7\% & 25.0\% \\

        & Matched (1.3)
        & 25.33 & 1455
        & 19.93 & 5337
        & 46.67 & 1727
        & 30.64 & 2840
        & 87.9\% & 32.5\% \\

        \addlinespace[3pt]

        \multirow{2}{*}{
            \shortstack[l]{
                \quad w/o Positive-Only\\
                \quad Gating
            }
        }
        & Original
        & 23.33 & 777
        & 16.87 & 3993
        & 39.60 & 1206
        & 26.60 & 1992
        & 53.0\% & 52.7\% \\

        & Matched (0.5)
        & 24.89 & 1468
        & 17.83 & 5249
        & 44.57 & 1680
        & 29.10 & 2799
        & 74.6\% & 33.5\% \\

        \addlinespace[3pt]

        \multirow{2}{*}{
            \shortstack[l]{
                \quad w/o One-Sided\\
                \quad Shaping
            }
        }
        & Original
        & 23.89 & 1174
        & 16.17 & 4096
        & 43.93 & 1475
        & 28.00 & 2248
        & 65.1\% & 46.6\% \\

        & Matched (0.6)
        & 24.11 & 1465
        & 18.43 & 5278
        & 47.80 & 1603
        & 30.11 & 2782
        & 83.4\% & 33.9\% \\

        \midrule
        \multicolumn{12}{l}{
            \textit{B. Adaptive scaling
            (structural constraints retained)}
        } \\
        \addlinespace[2pt]

        Fixed Overall Strength
        & $\lambda=0.1833$
        & 25.33 & 1561
        & 17.93 & 5487
        & 46.60 & 1762
        & 29.95 & 2937
        & 82.0\% & 30.2\% \\

        \quad Fixed $s_+$
        & $s_+=0.3234$
        & 25.56 & 1542
        & 18.40 & 5512
        & 48.30 & 1811
        & 30.75 & 2955
        & 88.9\% & 29.8\% \\

        \quad Fixed $s_{\mathrm{all}}$
        & $s_{\mathrm{all}}=0.4672$
        & 25.11 & 1577
        & 18.03 & 5509
        & 47.67 & 1793
        & 30.27 & 2960
        & 84.7\% & 29.6\% \\

        \bottomrule
    \end{tabular}%
}
\end{table*}

\paragraph{Structural ablations and compression matching.}
We remove QGLAS's three structural constraints either jointly or
individually while retaining its adaptive strength rule.
The joint ablation applies two-sided length shaping at the reward
level to all responses and uses the full-group mean length as its
reference. The adaptive statistics are still computed from the original
quality rewards, including the quality-favored subset used to define
$s_+$.

Removing these constraints can substantially change compression,
confounding direct quality comparisons at the original strength.
For all four structural ablations, we therefore vary only the scalar
shaping strength and select an operating point whose macro-average
response length is close to that of full QGLAS.
All other settings are held fixed, and evaluation quality is not used
to select the operating point.
Table~\ref{tab:full_ablation} reports both the original configurations
and the matched configurations used in Table~\ref{tab:ablation}.

\paragraph{Calibration of adaptive ablations.}
The adaptive ablations retain all structural constraints and replace
either the overall strength or one of its input statistics by a
constant.
Using stored rollout groups from steps 0--1000 of full QGLAS training
with $(\beta_{\min},\beta_{\max})=(0.3,0.6)$, we compute
\begin{equation}
\label{eq:ablation_fixed_statistics}
\begin{aligned}
\bar{\lambda}
&=
\frac{1}{|\mathcal{G}|}
\sum_{g\in\mathcal{G}} \lambda_g
= 0.1833, \\
\bar{s}_{+}
&=
\frac{1}{|\mathcal{G}|}
\sum_{g\in\mathcal{G}} s_{+,g}
= 0.3234, \\
\bar{s}_{\mathrm{all}}
&=
\frac{1}{|\mathcal{G}|}
\sum_{g\in\mathcal{G}} s_{\mathrm{all},g}
= 0.4672,
\end{aligned}
\end{equation}
where $\mathcal{G}$ denotes the stored rollout groups used for
calibration. Neither evaluation quality nor evaluation response length
is used to choose these constants.

Our GSPO implementation omits group-wise standard-deviation
normalization, so the full adaptive strength can be written as
$\lambda_g=F(s_{\mathrm{all},g},s_{+,g})$, where
\begin{equation}
\label{eq:ablation_strength_function}
F(s,t)
=
s\left[
\beta_{\min}
+(\beta_{\max}-\beta_{\min})
\left(
1-\operatorname{clip}
\left(\frac{t}{s+\epsilon},0,1\right)
\right)
\right].
\end{equation}
The three adaptive ablations use
\begin{equation}
\label{eq:ablation_adaptive_variants}
\lambda_g
=
\begin{cases}
\bar{\lambda},
& \text{Fixed overall strength}, \\[2pt]
F(s_{\mathrm{all},g},\bar{s}_{+}),
& \text{Fixed }s_{+}, \\[2pt]
F(\bar{s}_{\mathrm{all}},s_{+,g}),
& \text{Fixed }s_{\mathrm{all}}.
\end{cases}
\end{equation}
Thus, fixing $s_+$ removes its group-specific variation while retaining
the current group's overall reward spread.
Fixing $s_{\mathrm{all}}$ replaces both its outer multiplicative
factor and its occurrence in the ratio used to compute $\beta_g$.
All other QGLAS components remain unchanged, so every adaptive
ablation preserves the original advantage signs.

\paragraph{Results.}
Without compression matching, removing all structural constraints
increases CR from $32.4\%$ to $58.3\%$ and reduces QGR to $33.8\%$.
Removing positive-only gating or one-sided shaping also produces
substantially stronger compression, reaching $52.7\%$ and $46.6\%$ CR,
respectively.
After approximately matching compression, the joint structural ablation
retains only $60.0\%$ QGR at $31.2\%$ CR.
The individual structural ablations retain $74.6$--$87.9\%$ QGR,
all below full QGLAS's $102.0\%$.

Fixing the overall strength, $s_+$, or $s_{\mathrm{all}}$ yields
$82.0\%$, $88.9\%$, and $84.7\%$ QGR, respectively, despite weaker
compression of $29.6$--$30.2\%$.
Retaining either statistic's group-specific variation improves
quality retention over fixed overall strength in these comparisons,
but neither variant recovers the full method's performance.
Together, these results support complementary benefits from the
structural constraints and adaptation to both reward statistics.

\subsection{Optimization Interpretation and Pairwise Reordering}
\label{app:pairwise_ordering}

\begin{table}[t]
\centering
\small
\caption{
Pairwise ranking-reversal rates among positive-advantage responses.
``Overall'' is computed over all eligible pairs. Gap-quintile rows report
rates conditional on each 20\% bin of the pre-shaping quality-advantage
gap; smaller-gap bins correspond to more weakly separated quality
preferences.
}
\label{tab:positive_pair_reversal}
\begin{tabular}{lccc}
\toprule
Pair set
& Learned RM
& Rubric-based Judge
& LLM-as-a-Judge \\
\midrule
Overall
& 3.48\% & 3.02\% & 1.91\% \\
\midrule
Smallest-gap 20\%
& 12.30\% & 5.41\% & 5.09\% \\
20--40\%
& 4.02\% & 5.10\% & 2.57\% \\
40--60\%
& 0.95\% & 2.34\% & 0.69\% \\
60--80\%
& 0.14\% & 1.43\% & 0.73\% \\
Largest-gap 20\%
& 0.00\% & 0.83\% & 0.47\% \\
\bottomrule
\end{tabular}
\end{table}

Section~\ref{sec:positive_pair_reversal} examines whether QGLAS can override
clear quality preferences within the positive-advantage set. Here, we first
clarify the optimization interpretation of polarity preservation and then
characterize where QGLAS changes the relative ordering among quality-favored
responses.

\paragraph{Optimization interpretation.}
We study length control as a conservative intervention on top of a given
quality-only RL procedure. Our formulation is conditional on this procedure
providing a useful reference optimization: if the quality reward or the
quality-only training procedure is poorly specified, correcting that
optimization is a separate problem rather than a role of length control.
Under this premise, the role of the length signal is not to improve or correct
the quality objective, but to reduce generation length while perturbing the
reference learning behavior as little as possible. We therefore treat the sign
of the quality-induced advantage as the reference decision of whether a sampled
response should be reinforced or suppressed. Length may modulate the magnitude
of this signal, but should not reverse its polarity. Polarity preservation is
thus a fidelity constraint induced by our problem formulation, rather than a
claim that advantage sign is universally the uniquely correct invariant or that
preserving it guarantees unchanged downstream quality.

This constraint is intentionally weaker than preserving the complete ordering
of quality-induced advantages. A sign reversal changes whether a sampled
response is reinforced or suppressed relative to the quality-only reference,
whereas a ranking reversal within the positive-advantage set only reallocates
reinforcement strength among responses that remain quality-favored. We
therefore preserve the former while allowing the latter as a degree of freedom
through which conciseness can act. The targeted corrections in
Section~\ref{sec:flip_correction} provide complementary empirical
evidence that eliminating sign-conflicting updates improves quality retention
at similar compression. Below, we further examine whether the remaining
positive-set reorderings override clear quality preferences, and find that they
are concentrated among responses with small pre-shaping quality-advantage
gaps.

\paragraph{Pairwise reordering.}
To examine how QGLAS uses this remaining degree of freedom, we consider all
response pairs $(i,j)$ satisfying
\[
A_i^q > A_j^q > 0,
\]
so that response $i$ has a larger quality-induced advantage than response
$j$ before length shaping. We define the pre-shaping advantage gap as
\[
\Delta_{ij}^q = A_i^q - A_j^q
\]
and count a pairwise ranking reversal when
$\widetilde{A}_i < \widetilde{A}_j$.

Let $\mathcal{E}$ denote the set of all eligible pairs. The overall pairwise
reversal rate is
\[
R_{\mathrm{all}}
=
\frac{
\sum_{(i,j)\in\mathcal{E}}
\mathbb{I}[\widetilde{A}_i < \widetilde{A}_j]
}{
|\mathcal{E}|
}.
\]
To characterize how reversal frequency depends on the original quality
separation, we partition $\mathcal{E}$ into five equal-sized bins according
to $\Delta_{ij}^q$. For each gap quintile $\mathcal{B}_k$, we compute
\[
R_k
=
\frac{
\sum_{(i,j)\in\mathcal{B}_k}
\mathbb{I}[\widetilde{A}_i < \widetilde{A}_j]
}{
|\mathcal{B}_k|
}.
\]
Each reported quintile rate is therefore conditional on pairs within that
quintile.

As shown in Table~\ref{tab:positive_pair_reversal}, the overall reordering
rate within the positive-advantage set is low: 3.48\%, 3.02\%, and 1.91\%
under the Learned RM, Rubric-based Judge, and LLM-as-a-Judge settings,
respectively. More importantly, reordering is concentrated among pairs that
are weakly separated by the original quality signal.

For the Learned RM, the reversal rate decreases from 12.30\% in the
smallest-gap quintile to 0.95\% in the middle quintile and 0\% in the
largest-gap quintile. The same pattern holds under the other two reward
sources, for which fewer than 1\% of pairs in the largest-gap quintile are
reordered.

These results clarify the scope of QGLAS's polarity-preservation constraint.
QGLAS does not freeze the complete ordering induced by quality; instead, it
allows conciseness to redistribute reinforcement primarily among near-tied
quality-favored responses, while clearly separated quality preferences are
rarely overridden. In this sense, polarity preservation maintains fidelity
to the quality-only reference optimization, whereas adaptive magnitude
shaping provides the flexibility required for effective length control.

\begin{table}[t]
\centering
\caption{
Statistics of quality-favored responses per rollout group.
}
\label{tab:positive_subset}
\begin{tabular}{lcccc}
\toprule
Training Reward &
$|P|=0$ &
$|P|=1$ &
$|P|\ge4$ &
Mean $|P|$ \\
\midrule
Learned RM &
0.30\% &
0.10\% &
99.52\% &
8.16
\\
Rubric-based Judge &
0.08\% &
0.12\% &
99.02\% &
8.67
\\
LLM-as-a-Judge &
0.61\% &
0.20\% &
97.54\% &
8.46
\\
\bottomrule
\end{tabular}
\end{table}

\subsection{Distribution of Quality-Favored Responses}

QGLAS computes length shaping based on responses with positive
quality-induced advantages:
\[
P=\{i:A_i^q>0\}.
\]

When $P=\varnothing$, QGLAS skips length shaping. When $|P|=1$,
the reference length is identical to the only quality-favored response
length, resulting in zero length-shaping coefficient. Therefore,
effective length shaping requires multiple quality-favored responses.

To examine whether QGLAS typically operates with a sufficiently
populated quality-favored subset, we measure the distribution of
$|P|$ during training. As shown in Table~\ref{tab:positive_subset},
groups with no or only one quality-favored response are rare across
all reward sources. Moreover, more than 97\% of rollout groups contain
at least four positive-advantage responses, with an average of
8.16--8.67 positive responses out of 16.

\section{Evaluation Robustness}
\label{app:evaluation_robustness}

We conduct three complementary checks on the Arena-Hard-v2 evaluation:
direct head-to-head comparisons at matched compression, evaluation with
Arena-Hard-v2's built-in length control, and re-evaluation with an
alternative LLM judge.

\subsection{Direct Pairwise Evaluation at Matched Compression}
\label{app:direct_pairwise}

The main Arena-Hard-v2 evaluation scores each policy against a fixed
benchmark baseline. We additionally compare QGLAS directly against GR$^3$
and GRLC using the same Arena-Hard-v2 pairwise judging protocol. This
provides a direct test of whether QGLAS preserves higher response quality
at comparable compression.

We use the matched-compression Qwen3-4B configurations:
QGLAS $(\beta_{\min},\beta_{\max})=(0.3,0.6)$,
GR$^3$ with $\alpha=0.3$, and GRLC with
$(\lambda,\beta)=(0.2,0.2)$. Their average response lengths are closely
matched: 5379, 5296, and 5447 tokens on Hard Prompts, and 1723, 1774,
and 1668 tokens on Creative Writing, respectively.

We perform seed-matched comparisons for seeds 42, 43, and 44, pairing
QGLAS with the corresponding GR$^3$ or GRLC policy trained with the same
seed. For each prompt, the Arena-Hard-v2 protocol evaluates both answer
orders and aggregates the pairwise judgments into a score relative to the
baseline policy. A score of 50 is the indifference point; scores above 50
indicate an overall preference for QGLAS.

\begin{table}[t]
\centering
\small
\caption{
Direct Arena-Hard-v2 pairwise evaluation at matched compression.
Policies trained with the same seed are compared directly.
Scores above 50 indicate an overall preference for QGLAS.
The final column reports mean $\pm$ sample standard deviation across seeds.
}
\label{tab:direct_pairwise}
\begin{tabular}{lcccc}
\toprule
Comparison
& Seed 42
& Seed 43
& Seed 44
& Mean $\pm$ Std. \\
\midrule
\multicolumn{5}{l}{\textit{Hard Prompts}} \\
QGLAS vs.\ GR$^3$
& 57.3 & 59.2 & 59.7 & $58.7 \pm 1.3$ \\
QGLAS vs.\ GRLC
& 54.3 & 56.2 & 54.9 & $55.1 \pm 1.0$ \\
\midrule
\multicolumn{5}{l}{\textit{Creative Writing}} \\
QGLAS vs.\ GR$^3$
& 64.3 & 67.2 & 66.1 & $65.9 \pm 1.5$ \\
QGLAS vs.\ GRLC
& 71.3 & 68.8 & 70.3 & $70.1 \pm 1.3$ \\
\bottomrule
\end{tabular}
\end{table}

As shown in Table~\ref{tab:direct_pairwise}, QGLAS is preferred over both
baselines in every seed-matched comparison. Its mean scores against GR$^3$
are 58.7 on Hard Prompts and 65.9 on Creative Writing, while the
corresponding scores against GRLC are 55.1 and 70.1. All twelve
seed-by-subset comparisons are above the 50-point indifference level.
Thus, the quality advantage of QGLAS persists under direct head-to-head
evaluation at closely matched response lengths.

\subsection{Length-Controlled Arena-Hard Evaluation}
\label{app:length_controlled_evaluation}

Response length can confound preference-based evaluation because longer
answers may receive systematically different judgments from shorter ones.
This is particularly relevant here because NoBonus produces substantially
longer responses than the length-controlled policies.

We therefore use Arena-Hard-v2's built-in length-control procedure. Rather
than modifying responses or rerunning the judge, Arena-Hard-v2 adjusts the
aggregation of the same pairwise judgments. Specifically, its
Bradley--Terry model jointly fits model effects and a normalized
relative-response-length feature. The resulting model scores therefore
account for the systematic association between response length and
pairwise preference. This provides a controlled comparison while keeping
both the evaluated responses and the underlying judge decisions fixed.

\begin{table}[t]
\centering
\small
\caption{
Standard and length-controlled Arena-Hard-v2 evaluation on Qwen3-4B.
Length-controlled scores use Arena-Hard-v2's built-in length feature in
the pairwise Bradley--Terry aggregation. Best results among the
length-controlled RL methods are shown in bold.
}
\label{tab:length_controlled_eval}
\begin{tabular}{lcccc}
\toprule
& \multicolumn{2}{c}{Hard Prompts}
& \multicolumn{2}{c}{Creative Writing} \\
\cmidrule(lr){2-3}
\cmidrule(lr){4-5}
Method
& Standard $\uparrow$
& Length Ctrl. $\uparrow$
& Standard $\uparrow$
& Length Ctrl. $\uparrow$ \\
\midrule
Base
& 15.63 & 15.03
& 16.53 & 18.93 \\

NoBonus
& 21.03 & 16.60
& 48.20 & 46.20 \\

GR$^3$
& 17.07 & 14.07
& 45.47 & 44.47 \\

GRLC
& 20.13 & 16.93
& 42.50 & 41.17 \\

QGLAS
& \textbf{20.40} & \textbf{17.30}
& \textbf{50.47} & \textbf{48.33} \\
\bottomrule
\end{tabular}
\end{table}

Table~\ref{tab:length_controlled_eval} shows that the comparative result
persists after accounting for response length. On Hard Prompts, QGLAS
changes from 20.40 under standard evaluation to 17.30 after length control,
compared with 16.60 for NoBonus, 14.07 for GR$^3$, and 16.93 for GRLC.
On Creative Writing, QGLAS remains highest at 48.33, compared with 46.20
for NoBonus, 44.47 for GR$^3$, and 41.17 for GRLC.

The bootstrap intervals for QGLAS and NoBonus overlap: on Hard Prompts,
the length-controlled scores are $17.3_{-1.3}^{+1.1}$ and
$16.6_{-1.4}^{+1.4}$, respectively; on Creative Writing, they are
$48.3_{-2.6}^{+2.2}$ and $46.2_{-2.2}^{+2.5}$. We therefore do not
interpret the small QGLAS--NoBonus differences as statistically significant.
The relevant result is that, after accounting for response length, QGLAS
remains close to quality-only RL while outperforming both competing
length-control methods on both subsets.

\subsection{Robustness to the Choice of LLM Judge}
\label{app:judge_robustness}

Our main Arena-Hard-v2 evaluation uses GPT-4.1 as the judge. To test
sensitivity to judge choice, we re-evaluate the same model outputs using
Qwen3.5-397B-A17B-FP8. Only the judge model is changed; the evaluated
responses remain fixed.

\begin{table}[t]
\centering
\small
\caption{
Arena-Hard-v2 scores under two LLM judges. Values in parentheses denote
differences relative to NoBonus under the same judge. Despite different
absolute score scales, both judges produce the same ordering of the four
RL methods.
}
\label{tab:judge_robustness}
\begin{tabular}{lcccc}
\toprule
& \multicolumn{2}{c}{GPT-4.1}
& \multicolumn{2}{c}{Qwen3.5-397B-A17B-FP8} \\
\cmidrule(lr){2-3}
\cmidrule(lr){4-5}
Method
& Hard $\uparrow$
& Creative $\uparrow$
& Hard $\uparrow$
& Creative $\uparrow$ \\
\midrule
NoBonus
& 21.03 $(+0.00)$
& 48.20 $(+0.00)$
& 14.33 $(+0.00)$
& 59.87 $(+0.00)$ \\

GR$^3$
& 17.07 $(-3.96)$
& 45.47 $(-2.73)$
& 11.13 $(-3.20)$
& 56.40 $(-3.47)$ \\

GRLC
& 20.13 $(-0.90)$
& 42.50 $(-5.70)$
& 12.17 $(-2.16)$
& 54.20 $(-5.67)$ \\

QGLAS
& \textbf{20.40} $(-0.63)$
& \textbf{50.47} $(+2.27)$
& \textbf{14.10} $(-0.23)$
& \textbf{60.43} $(+0.56)$ \\
\bottomrule
\end{tabular}
\end{table}

As shown in Table~\ref{tab:judge_robustness}, the two judges differ
substantially in absolute score calibration but produce the same method
ordering. On Hard Prompts, both yield
NoBonus $>$ QGLAS $>$ GRLC $>$ GR$^3$; on Creative Writing, both yield
QGLAS $>$ NoBonus $>$ GR$^3$ $>$ GRLC. QGLAS also remains close to
NoBonus under both judges while outperforming GR$^3$ and GRLC on both
subsets.

Together, the direct pairwise comparisons, length-controlled aggregation,
and cross-judge evaluation provide complementary evidence that QGLAS's
quality advantage over the competing length-control methods persists across
evaluation protocols, after accounting for response length, and under an
alternative LLM judge.

\end{document}